\pdfoutput=1
\documentclass[times,twocolumn,final,authoryear]{elsarticle}

\usepackage{prletters}
\usepackage{framed,multirow}

\usepackage{amssymb}
\usepackage{latexsym}
\usepackage[inline]{enumitem}
\usepackage{multirow}
\usepackage{rotating}
\usepackage[utf8]{inputenc}
\usepackage{mathtools}
\usepackage{lineno}

\usepackage{url}
\usepackage{xcolor}
\usepackage{amsmath}
\usepackage{algorithm}
\usepackage{algpseudocode}

\usepackage{tikz}
\usetikzlibrary{decorations.pathmorphing}

\newcommand\CONDITION[2]%
  {\begin{tabular}[t]{@{}l@{}l@{}}
     #1&#2
   \end{tabular}%
  }
\algdef{SE}[WHILE]{While}{EndWhile}[1]%
  {\algorithmicwhile\ \CONDITION{#1}{\ \algorithmicdo}}%
  {\algorithmicend\ \algorithmicwhile}
\algdef{SE}[FOR]{For}{EndFor}[1]%
  {\algorithmicfor\ \CONDITION{#1}{\ \algorithmicdo}}%
  {\algorithmicend\ \algorithmicfor}
\algdef{S}[FOR]{ForAll}[1]%
  {\algorithmicforall\ \CONDITION{#1}{\ \algorithmicdo}}
\algdef{SE}[REPEAT]{Repeat}{Until}{\algorithmicrepeat}[1]%
  {\algorithmicuntil\ \CONDITION{#1}{}}
\algdef{SE}[IF]{If}{EndIf}[1]%
  {\algorithmicif\ \CONDITION{#1}{\ \algorithmicthen}}%
  {\algorithmicend\ \algorithmicif}%
\algdef{C}[IF]{IF}{ElsIf}[1]%
  {\algorithmicelse\ \algorithmicif\ \CONDITION{#1}{\ \algorithmicthen}}

\newcommand*{\rom}[1]{\expandafter\@slowromancap\romannumeral #1@}

\journal{Pattern Recognition Letters}

\begin{document}

\thispagestyle{empty}

\clearpage

\ifpreprint
  \setcounter{page}{1}
\else
  \setcounter{page}{1}
\fi

\begin{frontmatter}

\title{Strict Very Fast Decision Tree: a memory conservative algorithm for data stream mining}

\author[1]{Victor Guilherme Turrisi da Costa\corref{cor1}*} 
\ead{victorturrisi@uel.br}

\author[2]{André Carlos Ponce de Leon Ferreira de Carvalho\corref{cor2}}
\ead{andre@icmc.usp.br}

\author[1]{Sylvio Barbon Junior\corref{cor1}}
\ead{barbon@uel.br}

\address[1]{Computer Science Department, State University of Londrina, Londrina - PR, 86057-970, Brazil}

\address[2]{Institute of Mathematical and Computer Sciences, University of S{\~a}o Paulo - USP,  S{\~a}o Carlos - SP, 13566-590, Brazil}


\begin{abstract}
Dealing with memory and time constraints are current challenges when learning from data streams with a massive amount of data. Many algorithms have been proposed to handle these difficulties, among them, the Very Fast Decision Tree (VFDT) algorithm. Although the VFDT has been widely used in data stream mining, in the last years, several authors have suggested modifications to increase its performance, putting aside memory concerns by proposing memory-costly solutions.
Besides, most data stream mining solutions have been centred around ensembles, which combine the memory costs of their weak learners, usually VFDTs.
To reduce the memory cost, keeping the predictive performance, this study proposes the Strict VFDT (SVFDT), a novel algorithm based on the VFDT.
The SVFDT algorithm minimises unnecessary tree growth, substantially reducing memory usage and keeping competitive predictive performance.
Moreover, since it creates much more shallow trees than VFDT, SVFDT can achieve a shorter processing time.
Experiments were carried out comparing the SVFDT with the VFDT in 11 benchmark data stream datasets.
This comparison assessed the trade-off between accuracy, memory, and processing time.
Statistical analysis showed that the proposed algorithm obtained similar predictive performance and significantly reduced processing time and memory use.
Thus, SVFDT is a suitable option for data stream mining with memory and time limitations, recommended as a weak learner in ensemble-based solutions.

\end{abstract}


\end{frontmatter}

\section{Introduction}

Traditional machine learning (ML) algorithms work by modelling knowledge from static and previously collected datasets.
Currently, there is a growing demand for ML-based solutions able to deal with very large volumes of data, which usually comes in the form of continuous streams, creating new challenges.
Differently to learning from static data, which assumes that all training data necessary to induce a model is available, learning from data streams assumes that new data can arrive at any time, which can make a model outdated.
This may happen due to the occurrence of concept drifts, which are related to the change of data distribution in the problem space over time.
Therefore, learning from data streams requires continuous model updates.
An additional challenge posed by learning from data streams is the demand to perform accurate predictions at any time \citep{Krawczyk2017, Gama2010}.
Also, since the model updating must be fast and the memory available can be limited, depending on where it occurs, it is expected that a good algorithm is capable of efficiently dealing with processing time and memory space.

Many learning algorithms have been proposed to cope with some of these aspects.
Among them, the Very Fast Decision Tree (VFDT) algorithm \citep{Domingos2000} is one of the most well-known for stream classification, being capable of constructing a decision tree in an online fashion by taking advantage of a statistical property called Hoeffding Bound (HB).
By doing so, the VFDT obtains a predictive performance similar to conventional decision tree induction algorithms applied to static datasets \nocite{Domingos2000}.
Although VFDT is somewhat memory-friendly, learning from data streams can lead to unnecessary tree growth, increasing memory usage and even compromising its application on memory-scarce scenarios.

In the last years, \citep{Holmes2005, Yang2011-o-retorno, Yang2013} proposed a series of modifications to increase the predictive performance of the VFDT algorithm.
However, this came with a substantial increase in the memory cost.
Moreover, according to \cite{Krawczyk2017}, data stream researchers are shifting their focus to ensemble-based solutions.
The performance of these solutions depend on the strength of their base learners and the statistical correlation between them.
Hence, ensembles can use only weak learners as long as their correlation is low \citep{Breiman2001}.
Thus, learners with very similar predictive performance could be used as base learners for an ensemble and have virtually the same performance.
However, the use of several base-learners increase memory costs, limiting the use of ensembles.

In order to deal with memory cost restrictions, keeping the predictive performance, we propose a new base learner, called Strict Very Fast Decision Tree (SVFDT).
Our algorithm addresses these requirements, while being faster than the VFDT in some cases.
Thus, SVFDT can cope with memory-scare scenarios and ensemble-based solutions in the following way:
\begin{enumerate}
    \item SVFDT uses significantly less memory in comparison to VFDT, reaching similar predictive performance;
    \item SVFDT and VFDT were compared with various benchmark datasets through critical result analysis; 
    \item Two SVFDT versions were proposed, one designed to consume less memory and training time (SVFDT-I) and another (SVFDT-II) with a higher predictive performance.
\end{enumerate}

Experiments were performed on various benchmark datasets, measuring the accuracy, Kappa M, memory, and training time of SVFDT and VFDT, and performing a statistical test to assess significant statistical differences.

This paper is organised as follows.
Section 2 describes the VFDT algorithm.
Section 3 presents other ML algorithms for data streams similar to the VFDT.
Section 4 introduces the SVFDT, along with its pseudocode. Section 5 has an empirical study, comparing the proposed algorithm with VFDT, and discussing the results obtained.
Finally, Section 6 covers the conclusion and future work.

\section{Very Fast Decision Tree}

VFDT \citep{Domingos2000} is a tree-based ML algorithm for data streams designed around the principles of the HB.
The HB theorem states the following.
Suppose a continuous variable $v$, whose values are bounded by the interval $[v_{min}, v_{max}]$, with a range of values $R = v_{max} - v_{min}$.
Additionally, presume that this variable was independently observed $n$ times and the computed mean, according to these observations, is $\overline{v}$.
Thus, the HB theorem states that this variable has a true mean $\overline{v_{true}}$ (when $n\rightarrow \infty $) bounded by the interval $[\overline{v} - \epsilon, \overline{v} + \epsilon]$ with statistical probability $1 - \delta$, where

\begin{equation}\label{eq:hb}
    \epsilon = \sqrt{\frac{R\textsuperscript{2}\,ln(\frac{1}{\delta})}{2\,n}}.
\end{equation}

The VFDT algorithm applies the HB to evaluate if a given leaf should be split during the training phase.
After ranking split candidate features during a split attempt, according to a heuristic measure $G(.)$, VFDT uses the HB to check if the best split candidate would remain the best, had the tree observed more instances. 
Assuming that the features with the highest and second highest $G(.)$ values are $X\textsubscript{b}$ and $X\textsubscript{sb}$, respectively, let $\Delta G = G(X\textsubscript{b}) - G(X\textsubscript{sb})$. If $\Delta G > \epsilon$, then $X\textsubscript{b}$ holds as the best, with probability $1 - \delta$.
The $G(.)$ estimates correlation or dependence between two quantities, using metrics, such as Information Gain (IG) or Gini Index (GI).

Based on these assumptions, the VFDT is able to learn from a single instance at a time using limited computational memory resources.
Additionally, under realistic assumptions, it has the same asymptotic performance as a decision tree produced by a standard batch algorithm \citep{Gama2010}.
It is also worth mentioning that the VFDT, unlike batch decision tree induction algorithms, is capable of predicting new instances at any time.

The first version of the VFDT only handled nominal features.
Afterwards, many estimators for continuous features were proposed. 
\cite{Pfahringer:2008:HNA:1786574.1786604} reviewed these estimators and observed that the Gaussian estimator is the least sensitive to hyperparameter value and induced the most accurate models, becoming the default estimator in recent works.

To avoid unnecessarily split condition analysis, this checking is only executed if the leaf has an impure class distribution, i.e., there is more than one class of instances that fell on a given leaf.
Likewise, with the same goal, this check is only performed after $n$ instances fell into that leaf since the last check.
\citep{Domingos2000}. 
The authors also introduced a tiebreak hyperparameter $\tau$ to support tree growth when $\Delta G$ is very low.
This is done by checking if $\Delta G < \epsilon < \tau$ is true, ignoring the HB condition \citep{Domingos2000}.
It must be observed that a high value of $\tau$ may lead to tree size explosion and even completely ignoring the HB condition, e.g., when learning from a stream with two classes, using $n = 200$ and $\delta = 10^{-7}$, if $t \geq 0.201$ than the HB condition will never be checked.

Later, to increase VFDT predictive performance, instead of using a traditional most common (MC) prediction at the leaves, a Naive Bayes (NB) or Adaptive Naive Bayes (ANB) algorithm can be employed \citep{Gama2003}.

\section{Related Work}

Several works proposed modifications to the VFDT algorithm.
The Genuine Tie Detection \citep{Holmes2005} has a mechanism to automatically choose $\tau$ during training.
Despite the VFDT simplification by removing one hyperparameter, there was a decrease in the predictive performance for most of the datasets used in the experiments.

In a similar work, \cite{Yang2011} proposed Optimised-VFDT (OVFDT), whose goal was also to increase accuracy avoiding tree size explosion, substituting $\tau$ by statistics about the HB.
OVFDT was compared with three algorithms: VFDT (with multiple $\tau$ values); Genuine Tie Detection; and Hoeffding Option Tree (HOT) \citep{Pfahringer2007}.
It must be observed that none of the compared algorithms try to reduce tree size.
When compared with VFDT with $\tau = 0.05$ (VFDT-0.05), OVFDT obtained a small accuracy improvement (3\%) at the cost of creating trees 2.4 times larger.

\citet{Yang2011-o-retorno, Yang2013} extended the OVFDT
adding statistical constraints related to leaf accuracy. 
When compared with VFDT-0.05, despite the small improvement in predictive performance, they always produced larger trees.

Other VFDT modification, the Concept-adapting Very Fast Decision Tree (CVFDT) algorithm \citep{Hulten2001}, keeps secondary trees in memory, constantly assessed to check if they outperform the original tree, allowing adaptation to concept drifts.
Also, CVFDT uses a sliding window to discard old instances.
In the absence of concept drifts, the additional memory costs to store secondary trees makes CVFDT less efficient than VFDT-0.05, as shown in \citep{Yang2011}.
In concept drift scenarios, CVFDT predictive performance is much lower than those of ensemble-based solutions  \citep{Krawczyk2017}.

Another algorithm based on VFDT, the Hoeffding option tree (HOT) \citep{Pfahringer2007}, includes option nodes, which makes an instance go down into multiple leaves.
An option node is essentially a split node with multiple conditions.
Thus, a new instance travels along all children nodes whose conditions are true.
HOT performs a prediction by averaging the weight of the predictions of all leaves reached.
This algorithm presented  predictive performance higher than VFDT, at the cost of significant memory increase.

All of these previous modifications to VFDT provided better predictive performance, at the cost of an increase in memory and processing time. Our proposal aims at reducing these drawbacks while keeping a competitive predictive performance. In this way, we evaluate our algorithm using VFDT as the baseline.

\section{Strict Very Fast Decision Tree}
This section describes the proposed algorithm, Strict Very Fast Decision Tree (SVFDT). SVFDT modifies VFDT by strongly controlling tree growth without degrading predictive performance.
We propose two versions of the SVFDT, the SVFDT-I and SVFDT-II. In both versions, the following assumptions hold:
\begin{enumerate}
    \item A leaf node should split only if there is a minimum uncertainty of class assumption associated with the instances, according to previous and current statistics;
    \item All leaf nodes should observe a similar number of instances to be turned into split nodes;
    \item The feature used for splitting should have a minimum relevance according to previous statistics.
\end{enumerate}

We strongly suggest Entropy (H) and Information Gain (IG) for the first and third assumptions.
Likewise, both metrics are also employed to evaluate split feature candidates.
However, different functions that work in an analogous way to IG or GI could also be applied.

To avoid unnecessary growth, the following function is adopted, using as an underlying concept the 3-$\sigma$ rule:

\begin{equation}
    \varphi(x, X) =
    \begin{dcases}
        \text{True},    & \text{if } x \geq \overline{X} - \sigma(X) \\ 
        \text{False},   & \text{otherwise}
    \end{dcases}
\end{equation}

Where $X$ is a set of observed values, $\overline{X}$ is their mean, $\sigma(X)$ is their standard deviation, and $x$ is a new observation.
We assume that $X$ follows a normal distribution.

Additionally, a leaf can satisfy the VFDT split conditions (according to the HB or tiebreak value) and still remain a leaf if SVFDT considers this split unnecessary.
When leaves satisfy the VFDT split condition, statistics corresponding to it are marked with an underscored \textit{satisfyVFDT}.

At each leaf $l$, the following constraints are employed every time there is a split attempt:
\begin{enumerate}
    \item $\varphi(H_l, \{H_{l_0}, H_{l_1}, ..., H_{l_L}\})$, where the former parameter is the current entropy of $l$ and the latter is a set of all entropies of all current leaves $L$ in the tree, including $l$ (Statement 1);
    
    \item $\varphi(H_l, \{H_{satisfyVFDT_0}, H_{satisfyVFDT_1}, ..., H_{satisfyVFDT_S}\})$, where the latter parameter corresponds to the entropies computed at all $S$ times that a leaf satisfied the VFDT split conditions (Statement 1);
    
    \item $\varphi(IG_{l}, \{IG_{satisfyVFDT_0}, IG_{satisfyVFDT_1}, ..., IG_{satisfyVFDT_S}\})$, where $IG_{l}$ is the IG of the best split feature at $l$ and the latter parameter is a set of the IGs computed all $S$ times that a leaf satisfied the VFDT split conditions (Statement 3);
    
    \item $n\textsubscript{\textit{l}} \geq \overline{\{n_{satisfyVFDT_0}, n_{satisfyVFDT_1}, ..., n_{satisfyVFDT_S}\}}$, where the former parameter corresponds to the number of elements seen at $l$ and the latter to average number of elements observed at all $S$ times that a leaf satisfied the VFDT split conditions (Statement 2).
\end{enumerate}

We did not apply the function $\varphi$ in the last constraint, since it is always possible to satisfy it by waiting for more instances to be assigned to a given leaf.
On the contrary, the other constraints are not so easily satisfied in the same way, which may cause deadlocks that even learning a large amount of instances would not resolve.

Additionally to the $\varphi$ function, SVFDT-II has a skipping mechanism to speed-up growing by ignoring all previously presented constraints using the following function:

\begin{equation}
    \varpi(x, X) =
    \begin{dcases}
        \text{True},    & \text{if } x \geq \overline{X} + \sigma(X) \\ 
        \text{False},   & \text{otherwise}
    \end{dcases}
\end{equation}

At a split attempt, if either $\varpi(H_l, \{H_{split_0}, H_{split_1}, ..., H_{split_S}\})$ or $\varpi(IG_{l}, \{IG_{split_0}, IG_{split_1}, ..., IG_{split_S}\})$ hold true, then all the other $\varphi$ constraints are ignored.

The memory costs added to VFDT to compute the constraints 2, 3 and 4 are $O(1)$.
Complementary, the memory cost of constraint 1 is $O(L_{max})$, with $L_{max}$ being the maximum number of leaves observed during the tree induction.

Regarding time complexity, the first constraint has a cost of $O(L_{max})$, while the others have $O(1)$ complexity.
These costs corresponds to a single operation and so, the time complexity added to the whole induction process are $O(t_{satisfiedVFDT} * L_{max})$ and $O(t_{satisfiedVFDT})$ , respectively, where $t_{satisfiedVFDT}$ is the number of times a leaf satisfied the VFDT split conditions.
For SVFDT-II, we have an additional time cost of $O(t_{satisfiedVFDT})$ for each mechanism.
Although we have these additional costs, tree size is significantly reduced, making SVFDT training faster or similarly to the VFDT.

Algorithm \ref{alg:svfdt} shows the pseudocode of SVFDT. $H\textsubscript{statistics}$, $IG\textsubscript{statistics}$, $n\textsubscript{statistics}$ and $LH$ correspond to the additional statistics that are used to validate $\varphi(.)$ and $\varpi(.)$ operations.
Algorithm \ref{alg:cansplit} implements the function that checks whether a given leaf should be split.
In addition to the VFDT split check, we added the variables $\varrho$, $\xi$, $\kappa$ and $\psi$ to denote constraints 1, 2, 3, and 4, respectively.
It is worth reminding that all statistics are updated when the VFDT split conditions are satisfied.
The procedure of feature selection, invoked in line 30 of Algorithm~\ref{alg:svfdt}, remained like in \citep{Domingos2000}.

\begin{algorithm}[!t]

\caption{The SVFDT algorithm.}\label{alg:svfdt}

\begin{algorithmic}[1]
\scriptsize
\Require
\Statex $S$: the stream of instances
\Statex $GP$: the grace period
\Statex $\delta$: the error probability 
\Statex $\tau$: the tiebreak value
\Ensure
\Statex \textit{SVFDT}: a trained Strict Very Fast Decision Tree
\Statex

\Procedure{SVFDT}{$S, GP, \delta, \tau$}
\State Let SVFDT $\gets l_{root}$ \Comment{The root}
\State Initiate $H_{statistics}$, $IG_{statistics}$ and $n_{statistics}$ for $\varphi$ and $\varpi$ equations

\State Let $LH$ be the hash of leaves 

\State Let $n_{l_{root}} \gets 0$ \Comment{Number of elements seen at $l\textsubscript{root}$}
\State Let $LC_{l_{root}} \gets 0$ \Comment{Number of elements on last split check at $l_{root}$}
\State Let $F_{l_{root}} \gets \emptyset $ \Comment{Set of features removed from comparison}

\For {($(X, y)$ in $S$)} \Comment{X is the feature vector of an instance of class y in $S$}
\State Sort $(X, y)$ to its leaf $l$
\State Let $\hat{y} \gets$ prediction of $l$

\State Let $n_{l} \gets n_{l} + 1 $
\State Update feature estimators and class distribution at $l$ according to $(X, y)$
\If{(class distribution at $l$ is impure $\land$ $n_{l} - LC_{l} > GP$)}
\State Compute $HB$ and $IG(.)$ of features in $l\not\in F_{l}$
\State Let $rank \gets$ Sorted $IG(.)$ computed

\If{(\textit{\textbf{CanSplit}}($rank, HB, \tau, l, LH, H_{statistics}, IG_{statistics}, n_{statistics}$))}
\State Remove leaf $l$ from $LH$
\State Replace leaf $l$ with a split node
\For{each branch of the split}
\State Let $l_{new} \gets$ new leaf
\State Initiate all the feature estimators on $l_{new}$
\State Let class distribution on $l_{new}$ $ \gets $ post-split distribution of $l_{new}$
\State Let $n_{l_{new}} \gets $ sum of class distribution on $l_{new}$
\State Let $LC_{l_{new}} \gets $ $n_{l_{new}}$
\State Let $F_{l_{new}} \gets \emptyset $
\State Add leaf $l_{new}$ to $LH$
\EndFor
\Else
\State Let $LC_{l} \gets n_{l}$
\State \textit{\textbf{FeatureSelection}}($rank, HB, F_{l}$)
\EndIf

\EndIf
\EndFor
\State \Return SVFDT
\EndProcedure
\end{algorithmic}
\end{algorithm}

\begin{algorithm}[!t]
\caption{The split check algorithm.}\label{alg:cansplit}

\begin{algorithmic}[1]
\scriptsize
\Require 
\Statex $rank$: sorted list of $IG(.)$ per feature
\Statex $HB$: the Hoeffding Bound value
\Statex $\tau$: tiebreak value
\Statex $l$: the current leaf node
\Statex $LH$: the hash of leaves
\Statex $H_{statistics}$: statistics about entropy values
\Statex $IG_{statistics}$: statistics about IG(.) values
\Statex $n_{statistics}$: statistics about the number of elements seen values
\Ensure 
\Statex Boolean value
\Statex

\Procedure{CanSplit}{$rank, HB, \tau, l, LH, H_{statistics}, IG_{statistics}, n_{statistics}$}

\State Let $IG_{best}$ and $IG_{second\_best}$ $\gets$ the highest and second highest $IG(.)$

\If{$(IG_{best} - IG_{second\_best} > HB \lor HB < \tau)$}

\State Compute $\overline{H}_{LH}$ and $\sigma(H_{LH})$ using $LH$

\State Compute $\overline{H}$ and $\sigma(H)$ using $H_{statistics}$

\State Compute $\overline{IG}$ and $\sigma(IG)$ using $IG_{statistics}$

\State Compute $\overline{n}$ and $\sigma(n)$ using $n_{statistics}$

\State Let $H_{l}$ and $n_{l} \gets$ entropy and number of elements seen at $l$

\State Update $H_{statistics}$, $IG_{statistics}$ and $n_{statistics}$ with $H_{l}$, $IG_{best}$ and $n_{l}$, respectively
\State Let $svfdt\_ii\_constraints \gets H_{l} \geq \overline{H} + \sigma(H) \land IG_{best} \geq \overline{IG} + \sigma(IG)$

\If{($svfdt\_ii\_constraints$)} \Comment{SVFDT-II version only}
\State \Return \textbf{True}
\EndIf

\State Let $\varrho \gets H_{l} \geq \overline{H}_{LH} - \sigma(H_{LH})$ \Comment{Constraint 1}

\State Let $\xi \gets H_{l} \geq \overline{H} - \sigma(H)$ \Comment{Constraint 2}

\State Let $\kappa \gets IG_{best} \geq \overline{IG} - \sigma(IG)$ \Comment{Constraint 3}

\State Let $\psi \gets n_{l} \geq \overline{n} - \sigma(n)$ \Comment{Constraint 4}

\State Let $svfdt\_constraints \gets \varrho \land \xi \land \kappa \land \psi$
\If{($svfdt\_constraints$)}
\State \Return \textbf{True}
\EndIf
\EndIf
\State \Return \textbf{False}
\EndProcedure
\end{algorithmic}
\end{algorithm}

\begin{figure}[!htb]
\begin{center}
\scalebox{0.45}{
\includegraphics{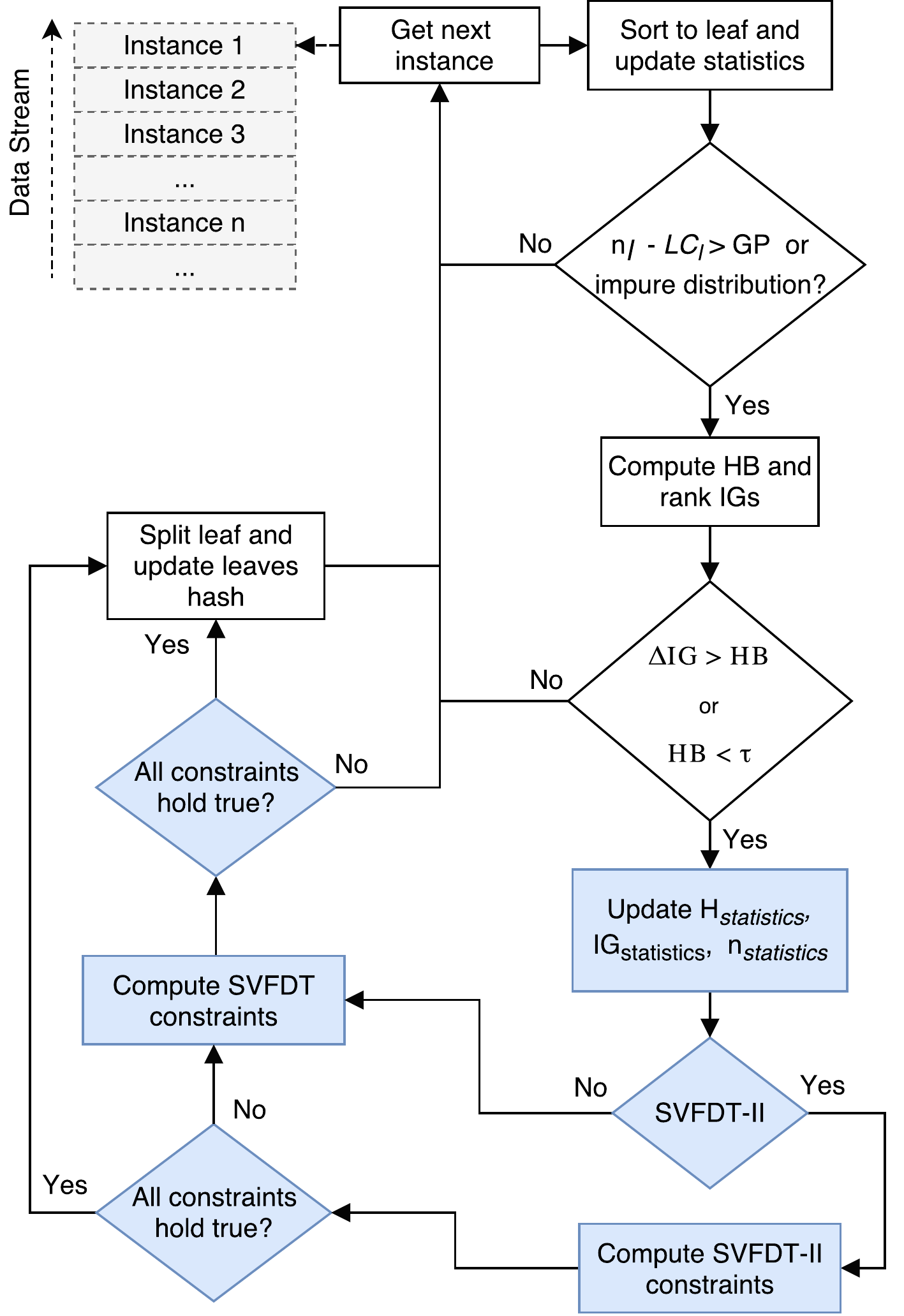}
}
\end{center}
\caption{
SVFDT diagram. Parts coloured in blue denote modifications in the traditional VFDT algorithm.
}
\label{fig:diagram-svfdt}
\end{figure}

Figure \ref{fig:diagram-svfdt} uses a flow chart to illustrate how the VFDT was modified to create the SVFDT-I and SVFDT-II (highlighted in blue).

\section{Empirical Study}
Both versions of SVFDT were experimentally compared with VFDT using 11 public datasets widely used in the data stream mining literature:
\begin{enumerate*}
\item Forest Cover Type dataset (covType) \citep{Moa};
\item Electricity Pricing dataset (elec) \citep{Moa};
\item Led datasets with 0\%, 10\% and 20\% noise composed of 1 million instances (led\_0, led\_10, and led\_20) \citep{Weka};
\item Random RBF datasets: $10^6$ instances with 10 features; 500,000 instances with 10 features; and 250 thousand instances with 50 instances (rbf\_1kk, rbf\_500k, and rbf\_250k(50)) \citep{Weka};
\item SEA dataset \citep{Street2001};
\item Spam dataset \citep{Katakis2010};
\item Usenet dataset \citep{Katakis2010}.
\end{enumerate*}
Table \ref{tab:summary_datasets} briefly describes these datasets.

\begin{table}[!htb]
\centering
\caption{Summary of the datasets used in the experiment.}
\label{tab:summary_datasets}
\scalebox{0.75}{
\begin{tabular}{|l|r|r|r|r|r|}
\hline
Dataset
& \begin{tabular}[c]{@{}c@{}}\# instances \end{tabular}           
& \begin{tabular}[c]{@{}c@{}}\# numeric\\ features \end{tabular} 
& \begin{tabular}[c]{@{}c@{}}\# binary\\ features\end{tabular} 
& \begin{tabular}[c]{@{}c@{}}\# categorical\\ features\end{tabular} 
& \begin{tabular}[c]{@{}c@{}}\# classes\end{tabular}       \\ \hline
covType        & 581,012                    & 10                     & 44                    & 0                          & 7                   \\ \hline

elec          & 45,312                     & 6                      & 0                     & 1                          & 2                   \\ \hline

led\_0         & \multirow{3}{*}{1,000,000} & \multirow{3}{*}{0}     & \multirow{3}{*}{24}   & \multirow{3}{*}{0}         & \multirow{3}{*}{10} \\  \cline{1-1}

led\_10        &                            &                        &                       &                            &                     \\  \cline{1-1}

led\_20        &                            &                        &                       &                            &                     \\ \hline

rbf\_1kk       & 1,000,000                  & \multirow{2}{*}{10}    & \multirow{3}{*}{0}    & \multirow{3}{*}{0}         & \multirow{3}{*}{2}  \\  \cline{1-2}

rbf\_500k     & 500,000                    &                        &                       &                            &                     \\  \cline{1-3}

rbf\_250k (50)  & 250,000                    & 50                     &                       &                            &                     \\ \hline

sea            & 60,000                     & 3                      & 0                     & 0                          & 2                   \\ \hline

spam           & 9,324                      & 0                      & 39,917                & 0                          & 2                   \\ \hline

usenet         & 5930                       & 0                      & 658                   & 0                          & 2                   \\ \hline
\end{tabular}
}

\end{table}

For each dataset, accuracy and Kappa M \citep{Bifet2015} measures for the three algorithms were computed, together with the number of tree nodes created.
The Kappa M was proposed to deal with unbalanced datasets toward measuring how a classifier compares with another that always predicts the majority class.
In the experiments, training time was calculated as the average of 30 runs.
Hyperparameter values recommended in the literature were used. These values are shown in Table \ref{tab:setup}.

\begin{table}[!htpb]
\centering
\caption{Hyperparameter values.}
\label{tab:setup}
\scalebox{0.923}{
\begin{tabular}{|c|c|c|c|} \hline
GP & $\tau$                   & Numeric estimator                & $\delta$  \\ \hline
200          & (0.05, 0.10, 0.15, 0.20) & Gaussian with 100 bins & $10^{-5}$ \\ \hline
\end{tabular}
}
\end{table}

All algorithms were implemented in Python 3.6, more specifically, VFDT coding was based on MOA's \citep{Moa}.

Table \ref{tab:performance} presents the accuracy, Kappa M, tree size and average training time for the four $\tau$ values adopted.
First, it is possible to observe that the accuracy values for both versions of the SVFDT are very close to those of the VFDT.
Likewise, Kappa M values are also close, since they are directly related to accuracy.
Regarding the size of the induced trees, it is possible to see a significant discrepancy.
In none of the tests performed, the size of SVFDT trees was larger than those of the VFDT trees, with the largest reduction for the rbf\_1kk dataset, where the number of nodes decreased from 3194 to 128 (4\% of the original size). 
Excluding the led24\_0 dataset, which is very simple and produces very small trees by default, SVFDT largely reduced tree size when compared to VFDT.
Finally, one of the main concerns was to avoid impacting training time due to the computation of the new constraints.
Although reducing training time is not the focus of this work, in many cases, the smaller trees resulted in shorter training times.

\begin{table}[htb]
\centering
\caption{Performance of each algorithm.}
\label{tab:performance}
\scalebox{0.72}{
\begin{tabular}{|l|l|r|r|r|r|}
\hline
\multicolumn{1}{|c|}{Dataset}                   & Algorithm & ACC & Kappa M & Size (nodes) & Time in sec. (std)  \\ \hline

\multirow{3}{*}{covType}            & VFDT      & 0.763 & 0.537             & 536           & 167.849 (0.532)         \\
                                    & SVFDT-I   & 0.758 & 0.529             & 365           & 135.510 (0.662)             \\
                                    & SVFDT-II & 0.763 & 0.537            & 467           & 159.249 (1.739)             \\
\hline

\multirow{3}{*}{elec}               & VFDT      & 0.801 & 0.531     & 209          & 6.235 (0.025)              \\
                                    & SVFDT-I     & 0.799 & 0.526              & 78            & 5.594 (0.021)              \\
                                    & SVFDT-II & 0.804 & 0.538              & 126           & 6.528 (0.018)              \\
\hline

\multirow{3}{*}{led24\_0}           & VFDT      & 1.000 & 1.000              & 19            & 27.114 (0.312)             \\
                                    & SVFDT-I     & 1.000 & 1.000              & 19            & 30.386 (0.436)              \\
                                    & SVFDT-II & 1.000 & 1.000              & 19            & 30.217 (0.483)              \\
\hline

\multirow{3}{*}{led24\_10}          & VFDT      & 0.733 & 0.703              & 554            & 33.918 (0.412)              \\
                                    & SVFDT-I     & 0.730 & 0.700             & 90            & 34.268 (0.338)              \\
                                    & SVFDT-II & 0.731 & 0.701              & 255            & 35.861 (0.228)              \\
\hline

\multirow{3}{*}{led24\_20}          & VFDT      & 0.504 & 0.449             & 524            & 33.952 (0.463)              \\
                                    & SVFDT-I     & 0.500 & 0.444              & 132            & 34.228 (0.231)              \\
                                    & SVFDT-II & 0.504 & 0.449             & 235            & 35.679 (0.232)              \\
\hline

\multirow{3}{*}{rbf\_1kk}     & VFDT      & 0.922 & 0.833              & 3194          & 245.774 (0.850)             \\
                                    & SVFDT-I     & 0.900 & 0.784             & 128           & 169.698 (0.858)             \\
                                    & SVFDT-II & 0.909 & 0.804             & 864           & 259.997 (0.987)             \\
\hline

\multirow{3}{*}{rbf\_500k}    & VFDT      & 0.914 & 0.815              & 1746          & 128.887 (0.361)             \\
                                    & SVFDT-I     & 0.894 & 0.771              & 124           & 100.163 (0.385)             \\
                                    & SVFDT-II & 0.900 & 0.785              & 504           & 138.246 (0.571)             \\
\hline

\multirow{3}{*}{rbf\_250k(50)}& VFDT      & 0.990 & 0.980              & 455           & 146.665 (0.331)             \\
                                    & SVFDT-I     & 0.982 & 0.964              & 110          & 185.454 (0.650)             \\
                                    & SVFDT-II & 0.990 & 0.979             & 312           & 197.425 (0.583)             \\
\hline

\multirow{3}{*}{sea}                & VFDT      & 0.850 & 0.598              & 273           & 6.409 (0.022)             \\
                                    & SVFDT-I     & 0.852 & 0.603              & 116           & 6.438 (0.019)              \\
                                    & SVFDT-II & 0.851 & 0.601              & 137           & 6.681 (0.019)              \\
\hline

\multirow{3}{*}{spam}               & VFDT      & 0.807 & 0.252             & 20            & 182.019 (1.708)             \\
                                    & SVFDT-I     & 0.768 & 0.102             & 8             & 174.086 (1.173)            \\
                                    & SVFDT-II & 0.768 & 0.102              & 8             & 173.826 (1.938)            \\
\hline

\multirow{3}{*}{usenet}             & VFDT      & 0.547 & 0.086              & 33            & 3.819 (0.163)             \\
                                    & SVFDT-I     & 0.563 & 0.119             & 11             & 3.787 (0.047)             \\
                                    & SVFDT-II & 0.561 & 0.114              & 9             & 3.795 (0.096)             \\
\hline

\end{tabular}
}
\end{table}

Table \ref{tab:relative-performance-tiebreak} shows the relative average accuracy, Kappa M, size and time for each $\tau$ value.
Relative metrics are obtained by dividing the value obtained by each SVFDT algorithm by the value of VFDT.
It is possible to see that, independently from the $\tau$ value, accuracy and Kappa M values were very similar.
The highest variation occurred when using $\tau = 0.10$, when SVFDT-I and SVFDT-II predictive performances decreased 1.6\% regarding VFDT.
Considering memory cost, the size of the trees produced by the SVFDT-I were at most 48\% of the size of the trees produced by VFDT on average. 
Although the SVFDT-II produced larger trees than the SVFDT-I, they were at most 67\% of the size of the VFDT trees on average.
Regarding training time, the SVFDT-I was faster for $\tau = 0.05$ and $\tau = 0.10$, with very significant gains of around 13\% to 15\%.
In contrast with the cases where the VFDT was faster, SVFDT-I was at most 10\% slower for $\tau = 0.20$.
SVFDT-II was faster than the VFDT only for $\tau = 0.05$.

\begin{table}[htb]
\centering
\caption{Mean relative metrics against the VFDT for each tiebreak value.}
\label{tab:relative-performance-tiebreak}
\scalebox{0.825}{
\begin{tabular}{|l|l|r|r|r|r|}
\hline
$\tau$             & Algorithm & Rel. ACC & Rel. Kappa M & Rel. size & Rel. time \\ \hline
\multirow{2}{*}{0.05} & SVFDT-I     & 0.992 & 0.991            & 0.484             & 0.853        \\
                      & SVFDT-II & 1.000 & 1.006            & 0.669             & 0.971       \\ \hline
\multirow{2}{*}{0.10} & SVFDT-I     & 0.984 & 0.844            & 0.412             & 0.877        \\
                      & SVFDT-II & 0.984 & 0.825            & 0.566             & 1.043        \\ \hline
\multirow{2}{*}{0.15} & SVFDT-I     & 0.992 & 1.020            & 0.347             & 1.044        \\
                      & SVFDT-II & 0.997 & 1.043            & 0.529             & 1.125        \\ \hline
\multirow{2}{*}{0.20} & SVFDT-I     & 0.999 & 1.076            & 0.353             & 1.103        \\
                      & SVFDT-II & 1.002 & 1.084            & 0.504             & 1.148        \\ \hline
\end{tabular}
}
\end{table}

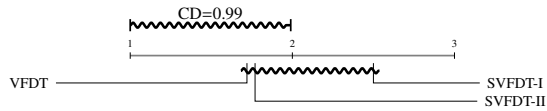
\begin{figure}[!htbp]
\centering 
\scalebox{0.8}{
\begin{tikzpicture}[xscale=2]
\node (Label) at  (01.9933,0.7) {\footnotesize{CD=0.99}}; 
\draw[decorate,decoration={snake,amplitude=.4mm,segment length=1.5mm,post length=0mm}, very thick, color = black](01.3333, 0.5) -- (02.6533, 0.5);
\foreach \x in {01.3333,02.6533} \draw[thick,color = black] (\x, 0.4) -- (\x, 0.6);

\draw[gray, thick](01.3333, 0) -- (04.0000, 0);
\foreach \x in {01.3333,02.6667,04.0000}\draw (\x cm,1.5pt) -- (\x cm, -1.5pt);
\node (Label) at (01.3333,0.2) {\tiny{1}};
\node (Label) at (02.6667,0.2) {\tiny{2}};
\node (Label) at (04.0000,0.2) {\tiny{3}};
\draw[decorate,decoration={snake,amplitude=.4mm,segment length=1.5mm,post length=0mm}, very thick, color = black](02.2433,-00.2500) -- ( 03.3833,-00.2500);
\node (Point) at (02.2933, 0){};  \node (Label) at (0.5,-00.4500){\scriptsize{VFDT}}; \draw (Point) |- (Label);
\node (Point) at (03.3333, 0){};  \node (Label) at (4.5,-00.4500){\scriptsize{SVFDT-I}}; \draw (Point) |- (Label);
\node (Point) at (02.3600, 0){};  \node (Label) at (4.5,-00.7500){\scriptsize{SVFDT-II}}; \draw (Point) |- (Label);
\end{tikzpicture}
}
\caption{Accuracy performance comparison among VFDT, SVFDT-I and SVFDT-II according to the Friedman and Nemenyi test using $\tau = 0.05$. There are no significantly different algorithms}
\label{nemenyi-acc}

\end{figure}

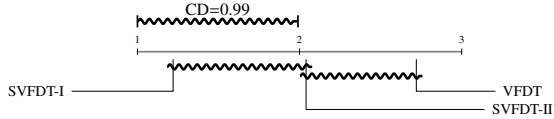
\begin{figure}[!htbp]
\centering 
\scalebox{0.8}{
\begin{tikzpicture}[xscale=2]
\node (Label) at  (01.9933,0.7) {\footnotesize{CD=0.99}}; 
\draw[decorate,decoration={snake,amplitude=.4mm,segment length=1.5mm,post length=0mm}, very thick, color = black](01.3333, 0.5) -- (02.6533, 0.5);
\foreach \x in {01.3333,02.6533} \draw[thick,color = black] (\x, 0.4) -- (\x, 0.6);

\draw[gray, thick](01.3333, 0) -- (04.0000, 0);
\foreach \x in {01.3333,02.6667,04.0000}\draw (\x cm,1.5pt) -- (\x cm, -1.5pt);
\node (Label) at (01.3333,0.2) {\tiny{1}};
\node (Label) at (02.6667,0.2) {\tiny{2}};
\node (Label) at (04.0000,0.2) {\tiny{3}};
\draw[decorate,decoration={snake,amplitude=.4mm,segment length=1.5mm,post length=0mm}, very thick, color = black](01.5767,-00.2500) -- ( 02.7700,-00.2500);
\draw[decorate,decoration={snake,amplitude=.4mm,segment length=1.5mm,post length=0mm}, very thick, color = black](02.6700,-00.4000) -- ( 03.6767,-00.4000);
\node (Point) at (01.6267, 0){};  \node (Label) at (0.5,-00.6500){\scriptsize{SVFDT-I}}; \draw (Point) |- (Label);
\node (Point) at (03.6267, 0){};  \node (Label) at (4.5,-00.6500){\scriptsize{VFDT}}; \draw (Point) |- (Label);
\node (Point) at (02.7200, 0){};  \node (Label) at (4.5,-00.9500){\scriptsize{SVFDT-II}}; \draw (Point) |- (Label);
\end{tikzpicture}
}
\caption{Memory performance comparison among VFDT, SVFDT-I and SVFDT-II according to the Friedman and Nemenyi test using $\tau = 0.05$. Algorithms that are not significantly different are connected}
\label{nemenyi-mem}
\end{figure}

\begin{figure}[!htbp]
\centering
\scalebox{0.8}{
\begin{tikzpicture}[xscale=2]
\node (Label) at  (01.9933,0.7) {\footnotesize{CD=0.99}}; 
\draw[decorate,decoration={snake,amplitude=.4mm,segment length=1.5mm,post length=0mm}, very thick, color = black](01.3333, 0.5) -- (02.6533, 0.5);
\foreach \x in {01.3333,02.6533} \draw[thick,color = black] (\x, 0.4) -- (\x, 0.6);

\draw[gray, thick](01.3333, 0) -- (04.0000, 0);
\foreach \x in {01.3333,02.6667,04.0000}\draw (\x cm,1.5pt) -- (\x cm, -1.5pt);
\node (Label) at (01.3333,0.2) {\tiny{1}};
\node (Label) at (02.6667,0.2) {\tiny{2}};
\node (Label) at (04.0000,0.2) {\tiny{3}};
\draw[decorate,decoration={snake,amplitude=.4mm,segment length=1.5mm,post length=0mm}, very thick, color = black](02.9767,-00.2500) -- ( 03.3167,-00.2500);
\node (Point) at (01.6933, 0){};  \node (Label) at (0.5,-00.4500){\scriptsize{SVFDT-I}}; \draw (Point) |- (Label);
\node (Point) at (03.2667, 0){};  \node (Label) at (4.5,-00.4500){\scriptsize{VFDT}}; \draw (Point) |- (Label);
\node (Point) at (03.0267, 0){};  \node (Label) at (4.5,-00.7500){\scriptsize{SVFDT-II}}; \draw (Point) |- (Label);
\end{tikzpicture}
}
\caption{Time performance comparison among VFDT, SVFDT-I and SVFDT-II according to the Friedman and Nemenyi test using $\tau = 0.05$. Algorithms that are not significantly different are connected}
\label{nemenyi-tempo}
\end{figure}
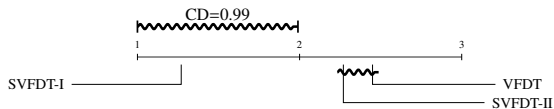

The statistical significance of the difference in accuracy, memory and training time were assessed using the Friedman's statistical test and the post-hoc test of Nemenyi.
A Critical Difference (CD) diagram is used to illustrate the results from these tests.
Figures \ref{nemenyi-acc}, \ref{nemenyi-mem} and \ref{nemenyi-tempo} present the CD diagrams for the accuracy, memory consumption and training time, respectively, using 95\% of significance.
They were constructed using the predictive performance of the trees with $\tau = 0.05$.
According to the statistical tests, there were no statistically significant differences between VFDT predictive performance and the predictive performance of the two proposed algorithms.
However, for memory used, there is a statistically significant difference only between SVFDT-I and VFDT, as shown in Figure \ref{nemenyi-mem}.
When considering training time, there was statistical difference between SVFDT-I and VFDT, but not
for SVFDT-II and VFDT.
Thus, for the datasets used in this study, SVFDT-I with $\tau = 0.05$ would be a better choice than VFDT, since it significantly reduced training time and memory, keeping a similar predictive performance.

\begin{figure*}[!htbp]
   \begin{center}
   \includegraphics[width=.33\textwidth]{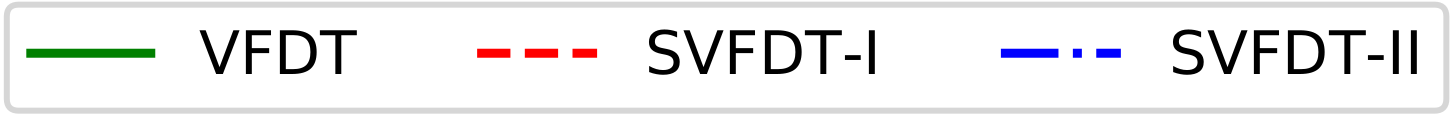}
   	\end{center}
	\setlength\tabcolsep{-1.8pt}
	\centering
	\begin{tabular}{cc}
        \vspace{-3pt}
	    \textbf{\scriptsize{covType}} &
        \textbf{\scriptsize{elec}} \\
	    \includegraphics[width=.505\textwidth]{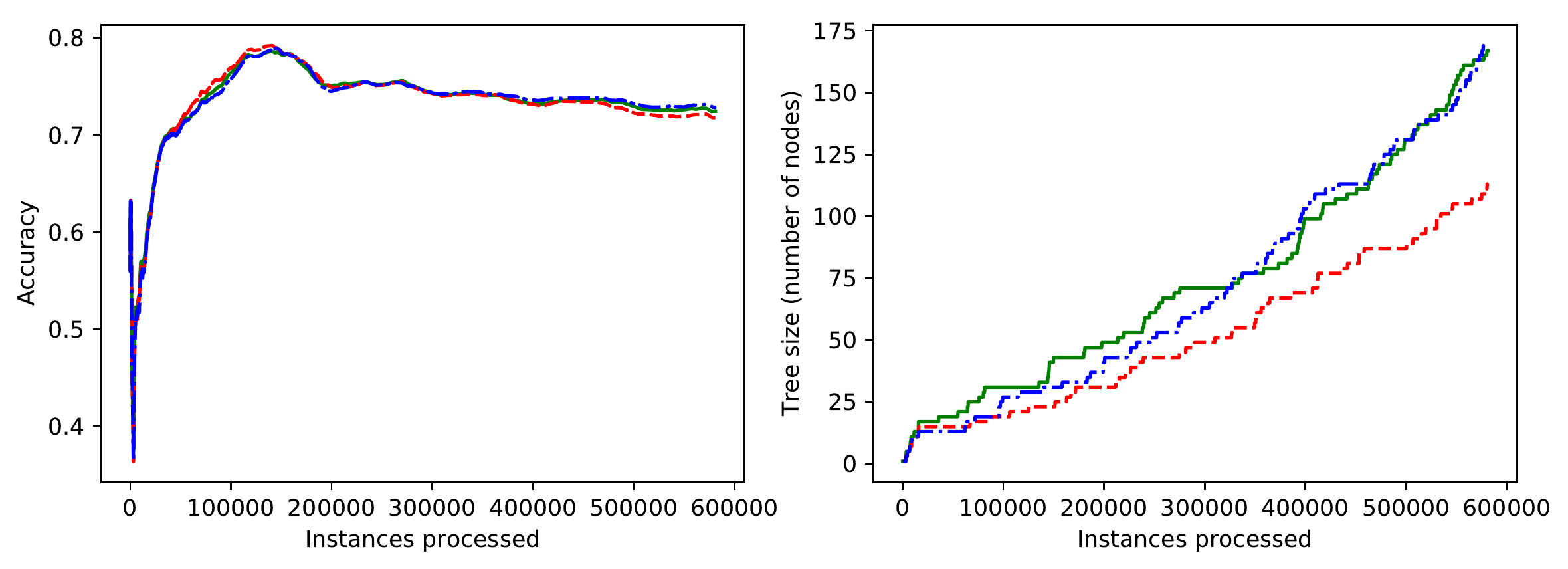} &
	    \includegraphics[width=.505\textwidth]{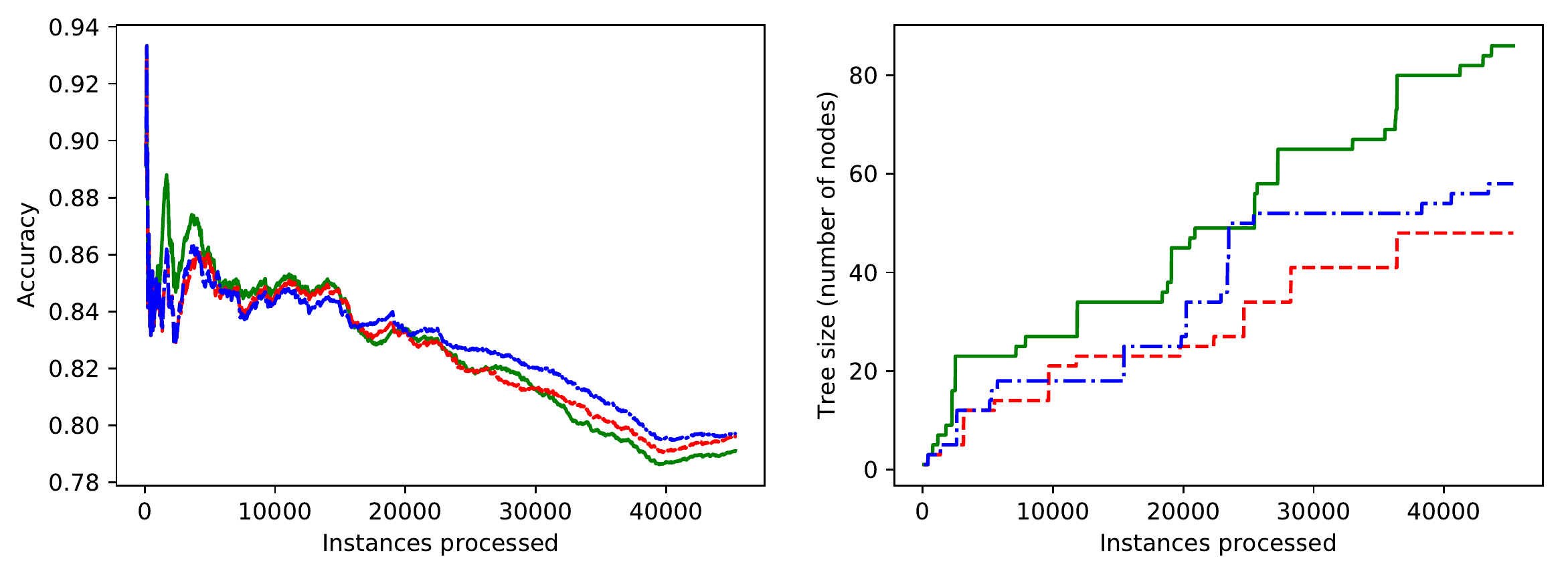} \\
	    
	    \vspace{-3pt}
	    \textbf{\scriptsize{led24\_0}} &
        \textbf{\scriptsize{led24\_10}} \\
	    \includegraphics[width=.505\textwidth]{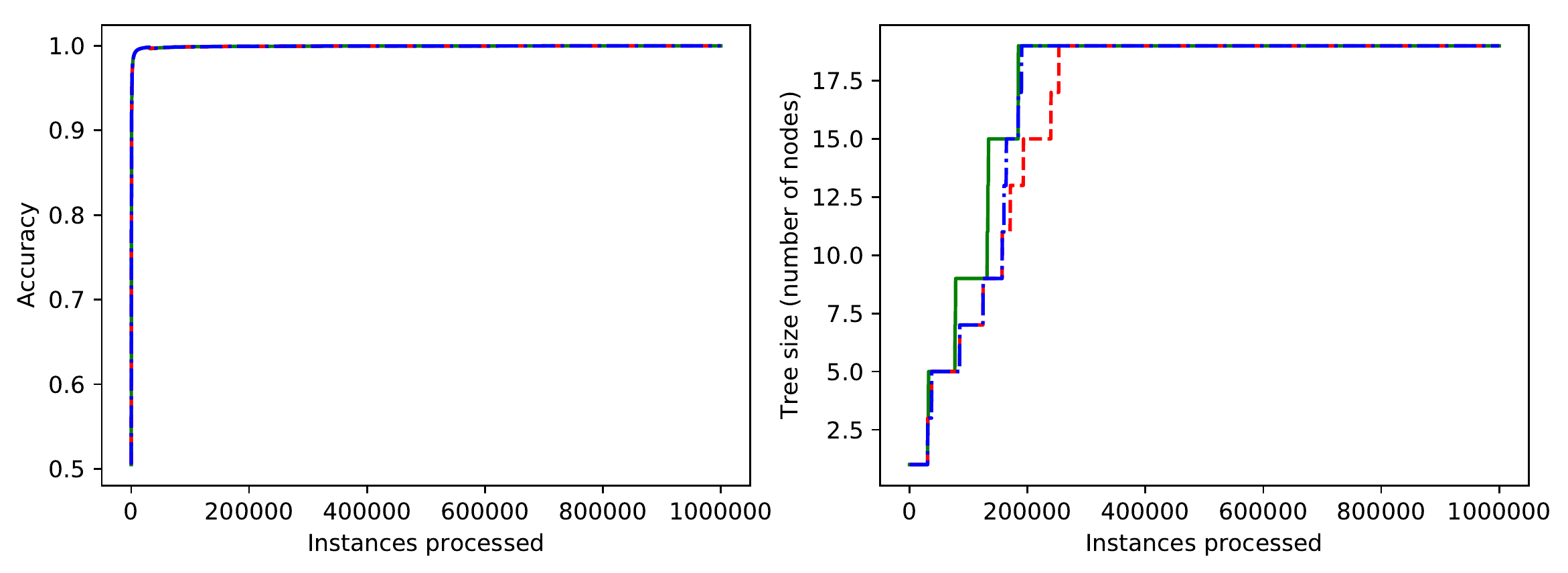} &
	    \includegraphics[width=.505\textwidth]{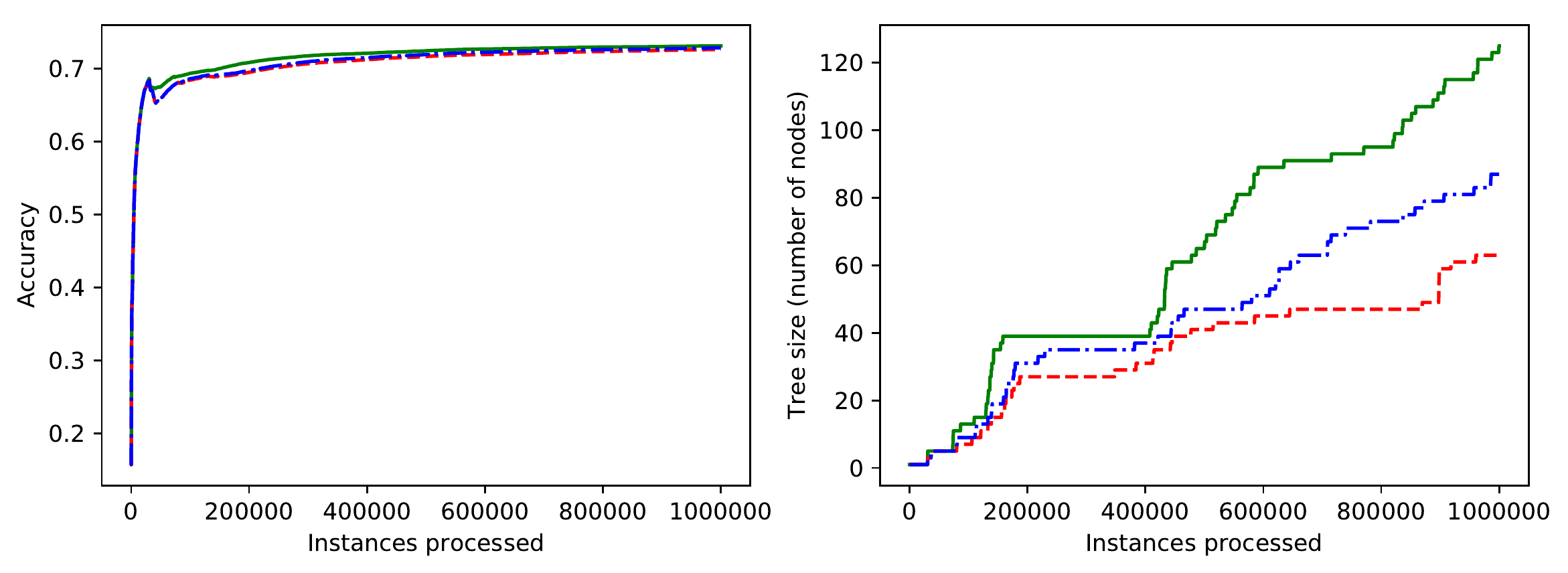} \\
	    
	    \vspace{-3pt}
	    \textbf{\scriptsize{led24\_20}} &
        \textbf{\scriptsize{rbf\_1kk}} \\
	    \includegraphics[width=.505\textwidth]{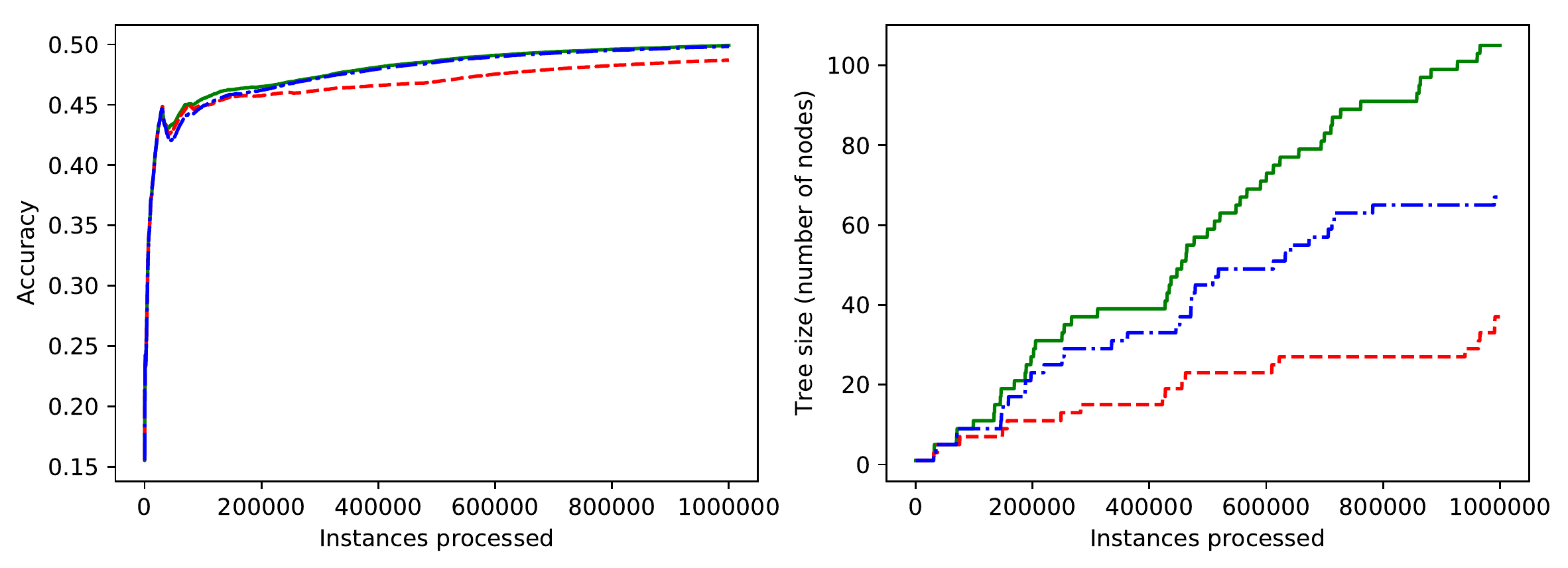} &
	    \includegraphics[width=.505\textwidth]{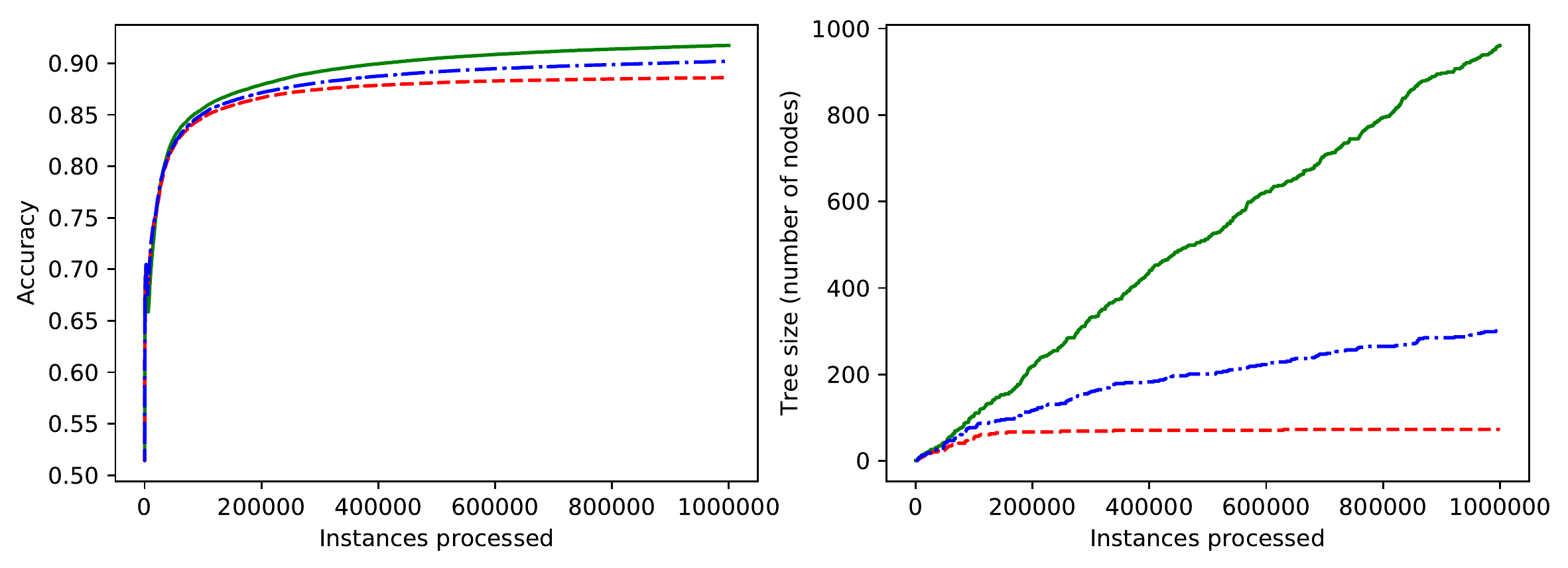} \\
	    
	    \vspace{-3pt}
	    \textbf{\scriptsize{rbf\_500k}} &
        \textbf{\scriptsize{rbf\_250k(50)}} \\
	    \includegraphics[width=.505\textwidth]{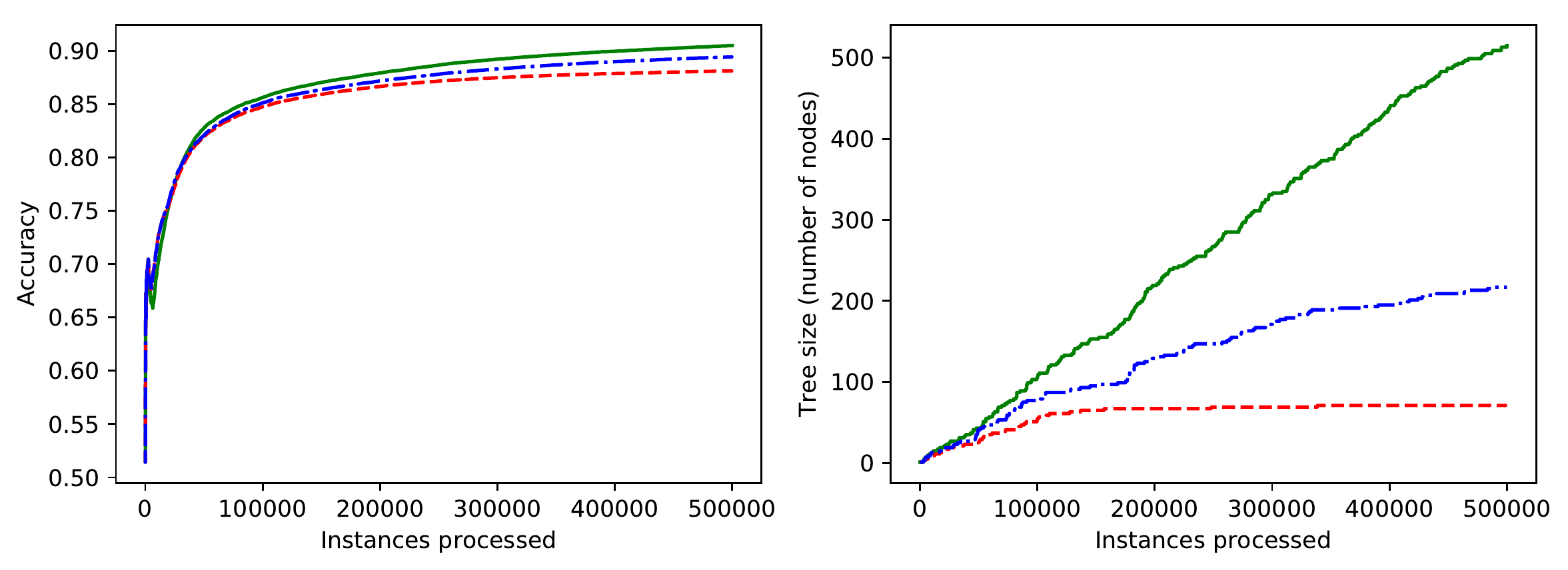} &
	    \includegraphics[width=.505\textwidth]{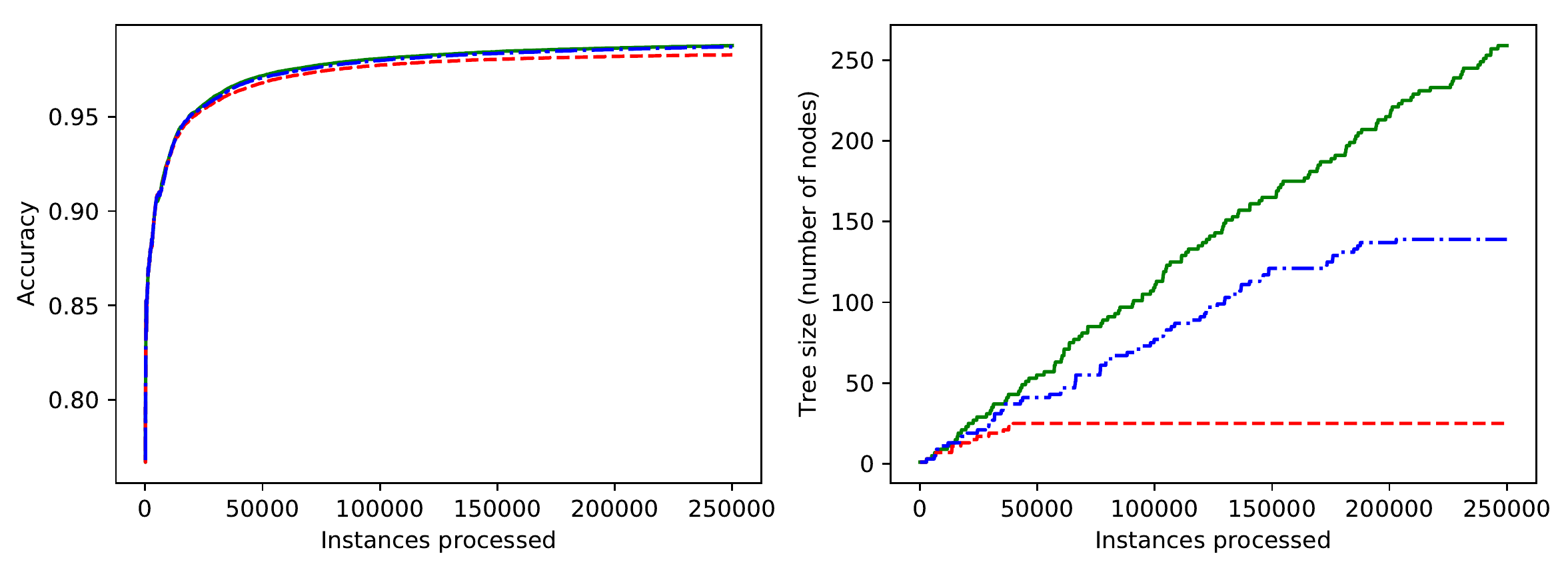} \\
	    
	    \vspace{-3pt}
	    \textbf{\scriptsize{sea}} &
        \textbf{\scriptsize{spam}} \\
	    \includegraphics[width=.505\textwidth]{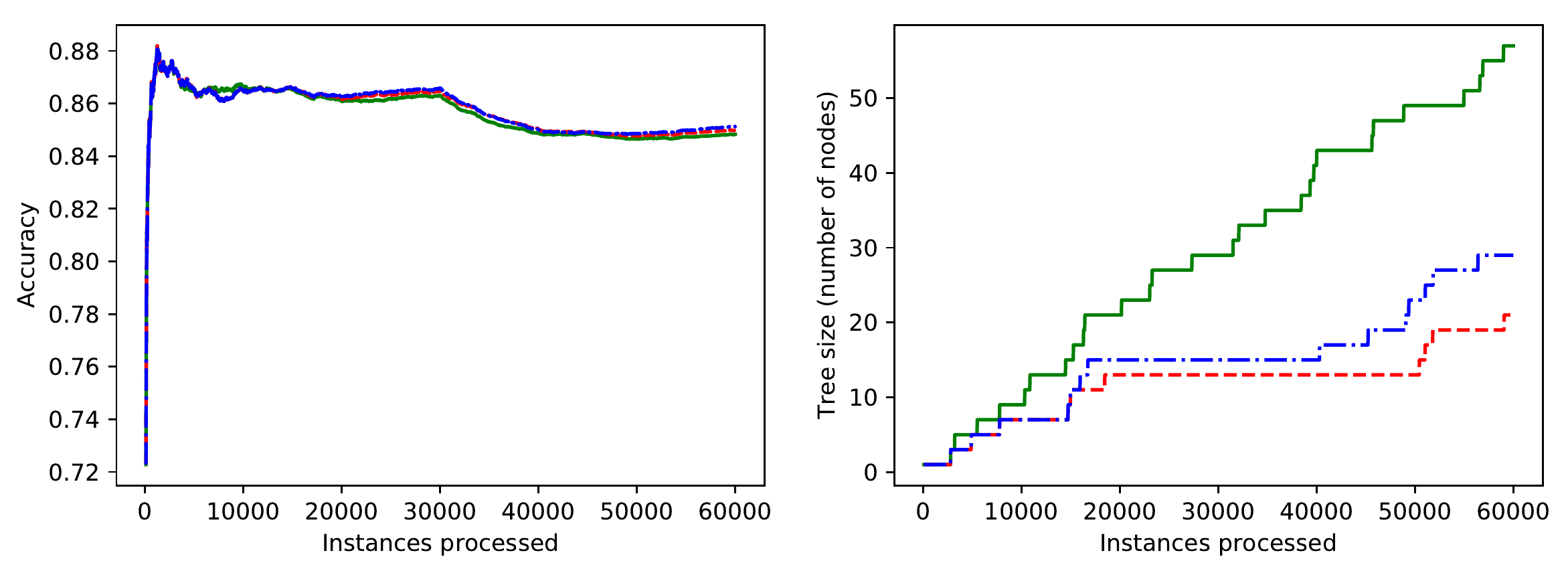} &
	    \includegraphics[width=.505\textwidth]{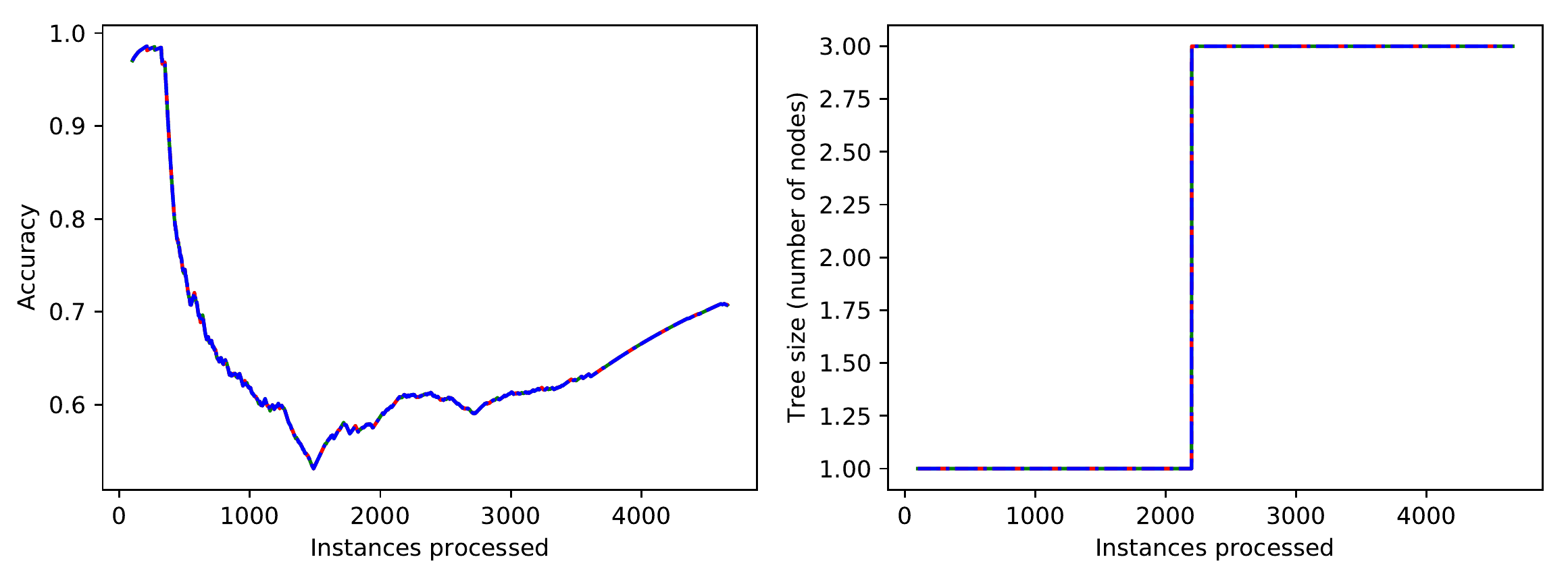} \\
	    
	\end{tabular}
    
    \centering
    \textbf{\scriptsize{usenet}} \\
    \includegraphics[width=.505\textwidth]{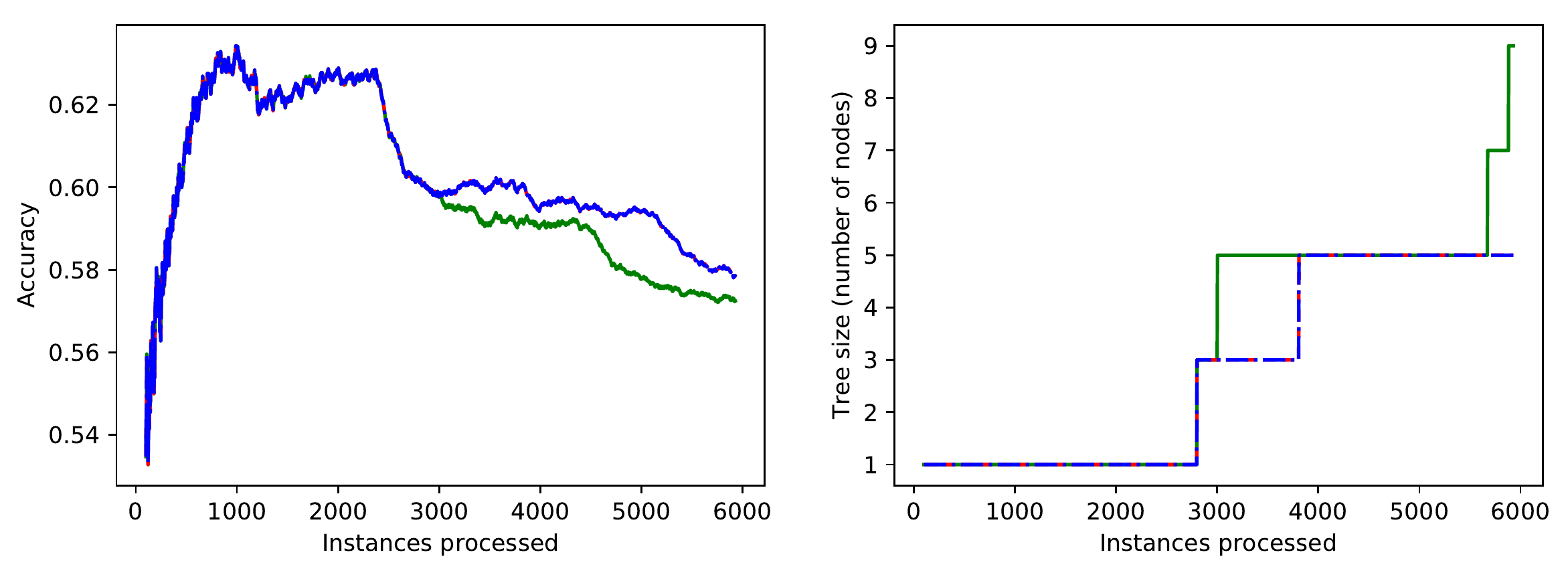}\\
    \caption{Accuracy and tree size (in number of nodes) for training/testing with a tiebreak of 0.05}
    \label{fig:acc-size-1}
\end{figure*}

The predictive performance of an algorithm on data stream classification can be evaluated by looking at its performance along the stream \citep{Gama2003}.
Accordingly, Figure \ref{fig:acc-size-1} presents the accuracy and tree size, in number of nodes, during training, per dataset.
It is possible to see that, as more instances were processed, both SVFDT algorithms kept predictive performance similar to VFDT.
But when considering the size of the trees, both SVFDTs outperformed VFDT by a large margin in most datasets.
It is also possible to notice periods where SVFDTs completely stops growing, while VFDT continues to grow, indicating that during these periods there is no need to increase the model size.
This pattern can be observed in all datasets.

Datasets with concept drift and noise were analysed.
Concept drifts are present in the sea, spam and usenet datasets.
SVFDTs and VFDT predictive performance in the presence of concept drifts were similar, except in the dataset sea, when SVFDTs' predictive performance was better, using less than half of the memory used by VFDT.

Performance in the presence of noise was explored in the led dataset, more specifically, led24\_10 and led24\_20 (Figure~\ref{fig:acc-size-1}).
In these datasets, the SVFDTs were still able to significantly reduce the tree size regarding VFDT.

\section{Conclusion and Future Work}

This work proposed and experimentally investigated two versions of a new VFDT-based algorithm, SVFDT-I and SVFDT-II.
The SVFDTs were created to reduce the size of the trees induced by VFDT, inducing a memory conservative decision tree for data stream mining. According to experimental results,
Both SVFDTs induce trees significantly smaller than those induced by the VFDT, while not statistically compromising predictive performance.
This study also assessed the influence of SVFDT-I hyperparameter $\tau$  value in the training time and size of induced trees.
For all $\tau$ values investigated, on average, trees at least 52\% smaller with accuracy at most 1.6\% lower than VFDT were created.
Sometimes this came with an increase in training time, showing that "there is no free lunch". 
Finally, for almost all datasets, SVFDT-II presented higher predictive accuracy than the SVFDT-I, together with significantly reducing tree size.
A statistical analysis of the performances of SVFDTs, when compared with VFDT, for $\tau = 0.05$, the value with the best results, showed no statistically significant difference in predictive performance, but a significant lower memory use and training time for SVFDT-I.
These results show that SVFDTs can be an efficient alternative to the VFDT in data stream mining applications.
As future work, we intend to investigate how to combine the proposed algorithms in ensembles, to increase predictive accuracy keeping low memory use and training time.

\bibliographystyle{model2-names}
\bibliography{refs}

\end{document}